\documentclass[letterpaper, 10 pt, conference]{ieeeconf}  

\IEEEoverridecommandlockouts                              

\usepackage{amsmath} 
\usepackage{color}
\usepackage{amsfonts}
\usepackage{bm}
\usepackage{booktabs}
\usepackage[T1]{fontenc}
\usepackage[pdftex]{graphicx}

\usepackage{caption}
\usepackage{url}
\let\labelindent\relax
\usepackage{enumitem}

\title{\LARGE \bf
Multi-goal Audio-visual Navigation using Sound Direction Map
}

\author{Haru Kondoh$^{1}$ and Asako Kanezaki$^{1}$
  \thanks{$^{1}$Haru Kondoh and Asako Kanezaki are with School of Computing, Department of Computer Science, Tokyo Institute of Technology, Japan}
}

\begin{document}

\maketitle
\thispagestyle{empty}
\pagestyle{empty}

\begin{abstract}

Over the past few years, there has been a great deal of research on navigation tasks in indoor environments using deep reinforcement learning agents.
Most of these tasks use only visual information in the form of first-person images to navigate to a single goal.
More recently, tasks that simultaneously use visual and auditory information to navigate to the sound source and even navigation tasks with multiple goals instead of one have been proposed.
However, there has been no proposal for a generalized navigation task combining these two types of tasks and using both visual and auditory information in a situation where multiple sound sources are goals.
In this paper, we propose a new framework for this generalized task: multi-goal audio-visual navigation.
We first define the task in detail, and then we investigate the difficulty of the multi-goal audio-visual navigation task relative to the current navigation tasks by conducting experiments in various situations.
The research shows that multi-goal audio-visual navigation has the difficulty of the implicit need to separate the sources of sound.
Next, to mitigate the difficulties in this new task, we propose a method named sound direction map (SDM), which dynamically localizes multiple sound sources in a learning-based manner while making use of past memories.
Experimental results show that the use of SDM significantly improves the performance of multiple baseline methods, regardless of the number of goals.

\end{abstract}

\section{INTRODUCTION}

Visual navigation tasks in indoor environments with deep reinforcement learning agents have been a research area of particular interest in the last decade.
Basic visual navigation uses only visual information in the form of first-person images to navigate to a single goal.
In recent years, more advanced tasks have emerged, such as audio-visual navigation~\cite{chen2020soundspaces}, which uses auditory as well as visual information to navigate to a sound source, and multi-object navigation (MultiON)~\cite{wani2020multion}, which navigates to not one but multiple goals. 
However, a task that uses both visual and auditory information to navigate to multiple sound source goals, i.e., a combination of audio-visual navigation and MultiON, has not yet been proposed.
In terms of real-world applications, there are many tasks, such as lifesaving or bird and animal control, where auditory information is helpful and there is not necessarily a single goal.

\begin{figure}[t]
    \begin{center}
        \centering
        \includegraphics[width=\linewidth]{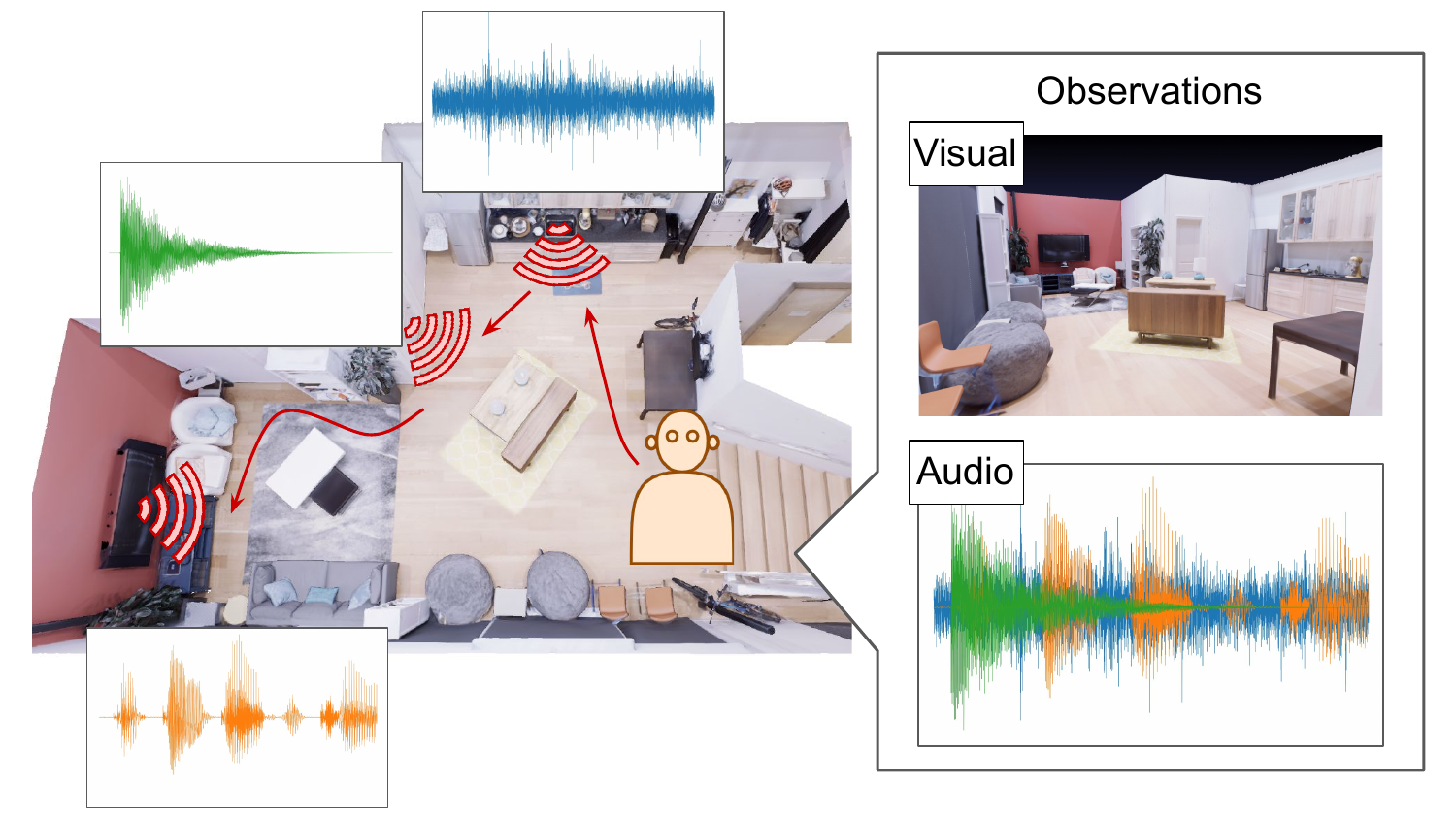}
        \caption{
            Overview of the multi-goal audio-visual navigation.
            Here, navigation is performed to three different sound sources in an indoor environment.
            The agent observes the first-person visual information and the auditory information, which is a superposition of sounds from three different sound sources. The agent must make appropriate action choices.
        }
        \label{fig:multi_goal_av_nav}
    \end{center}
\end{figure}

In this study, we propose a new task \textit{multi-goal audio-visual navigation} (Fig.~\ref{fig:multi_goal_av_nav}), which combines audio-visual navigation and MultiON.
There are three key elements to solving multi-goal audio-visual navigation:
sound source separation, memory, and action planning.
First, the reinforcement learning agent observes images and spectrograms of multiple overlapping sounds.
Therefore, accurate sound source separation plays an important role in improving the accuracy of sound source localization.
In addition, by remembering the information acquired before reaching one goal, it is expected to be able to efficiently navigate to the next goal.
Finally, action planning is important because it is necessary to infer which sound source should be the next goal to make an efficient route plan.

This study has two objectives.
The first objective is to identify where the difficulties in multi-goal audio-visual navigation lie by conducting experiments in various situations.
The second objective is to propose a new method for solving multi-goal audio-visual navigation with higher accuracy.
To overcome the difficulties found in this new task, we propose a method based on implicit dynamic multiple sound source localization, which is named a sound direction map (SDM).
The SDM aids path planning by simultaneously localizing multiple sound sources in a learning-based manner.
The SDM is dynamically updated by making effective use of memory.
This dynamic updating method can potentially improve the performance of sound source separation by utilizing previously predicted sound source localization information.

The three main contributions of this paper are summarized as follows.

\begin{itemize}[leftmargin=*]
\item A new navigation framework \textit{multi-goal audio-visual navigation} is proposed. We tested in a variety of situations to examine where the difficulties of this task lie.
\item As an efficient method for solving the new task, we propose the sound direction map (SDM), which represents the history of implicitly predicted dynamic multiple sound source locations.
\item In SoundSpaces 2.0~\cite{chen2022soundspaces}, we show that the proposed SDM consistently improves the performance of multiple baseline methods in all situations.
\end{itemize}
\section{RELATED WORK}

\subsection{Sound Source Localization}

Sound source localization is the estimation of the location of the sound source from the observed sound data \cite{rascon2017localization}.
In recent years, many methods based on deep learning have been proposed~\cite{grumiaux2022survey,ma2015exploiting,he2018deep}.
Some of these, such as Ma et al. \cite{ma2015exploiting} and He et al. \cite{he2018deep}, solve the problem of localizing multiple sound sources.
These methods are based on multilayer perceptron (MLP) and convolutional neural network (CNN) to estimate the direction of arrival of multiple sound sources.
Furthermore, Adavanne et al. \cite{adavanne2019localization} proposed a method using a convolutional recurrent neural network (CRNN) to track the direction of arrival of multiple sound sources in a situation where the sound sources are moving.

This paper also proposes a method for multiple sound source localization using deep learning.
In particular, we propose a method using reinforcement learning that dynamically localizes multiple sound sources while the observer moves.

\subsection{Visual Navigation}

Visual navigation is the process by which an autonomous mobile robot placed in an environment generates an appropriate and safe path from the start to a goal by using visual information \cite{bonin2008visual}.
In visual navigation in the three-dimensional indoor environment addressed in this study, first-person RGBD images are often used as the visual input.
Visual navigation requires the use of such visual information to understand the environment while estimating self and target positions and making appropriate action choices.

In recent years, many methods for solving visual navigation based on deep reinforcement learning have been proposed.
Parisotto et al. \cite{DBLP:journals/corr/ParisottoS17} proposed NeuralMap, a method to adaptively create a two-dimensional map to remember the spatial structure of the environment, based on the importance of memory for the partial observability of navigation.
Chaplot et al. \cite{chaplot2019learning, chaplot2020object} also proposed a method that simultaneously creates a map and estimates its own pose.
While these methods explicitly create maps, there are some methods that do not.
Fang et al. \cite{fang2019scene} and Fukushima et al. \cite{9812027} proposed a method that stores the embedding of observations as history and allows long-term memory consideration by using Transformer~\cite{vaswani2017attention}. 
A method is also proposed to improve navigation performance by simultaneously solving auxiliary tasks not directly related to navigation, such as RGB image depth prediction and closed-loop detection \cite{mirowski2017learning, 9370169}.

The closest prior work on visual navigation to this study is Wani et al. \cite{wani2020multion}, which proposed a task called multi-object navigation (MultiON).
In MultiON, navigation must be performed to multiple specified objects in a specified order.
In contrast, the order in which the goals should be reached is not determined in our multi-goal audio-visual navigation handled in this study.
It is therefore up to the agents themselves to decide which goal they want to achieve first.

\subsection{Audio-Visual Navigation}

In audio-visual navigation, the agent uses auditory as well as visual information to navigate to a sound source.
Auditory information may be useful for estimating self and target locations, which are important for navigation tasks, and for understanding the geometric structure of the environment.
Indeed, Chen et al. \cite{chen2020soundspaces} experimentally demonstrated the usefulness of using auditory information in navigation.

In recent years, this audio-visual navigation has been actively studied and a variety of tasks have been proposed.
Majumder and Pandey~\cite{majumdersemantic} proposed audio-visual navigation that uses a sound source other than the target as the interfering sound.
In addition, while conventional audio-visual navigation assumes that the sound never stops playing, Chen et al. \cite{chen2021semantic} proposed an audio-visual navigation framework in which the sound only plays for a short period immediately after the start of the episode.
Also, some proposed one in which the sound source moves \cite{younes2021catch, yu2021sound} and another realistic one in which the agent searches for a fallen object \cite{gan2022finding}.

Several methods for solving audio-visual navigation have already been proposed.
Chen et al. \cite{chen2020soundspaces} proposed AV-Nav, a simple method to obtain policy by inputting image features and spectrogram features into a gated recurrent unit (GRU)~\cite{chung2014empirical}. 
Then, Chen et al. \cite{chen2020learning} proposed AV-WaN, a method that uses the observed depth image and spectrogram to create a geographic map of the environment and an acoustic map that stores the sound intensity at that location to generate the next waypoint to be reached. 
Chen et al. \cite{chen2021semantic} proposed a Transformer-based policy network and a module that predicts the category and relative location of a sound source.
Furthermore, Tatiya et al. \cite{tatiya2022knowledge} proposed K-SAVEN, a knowledge-driven method that utilizes a knowledge graph, which represents object-to-object and object-to-region relationships, and extracts features from it using graph convolutional network \cite{kipf2017semisupervised}.

While a variety of tasks and methods have been studied, the task of navigating to multiple sound sources in a single episode has not been studied.
Therefore, this study is the first to address this task.

For the performance evaluation, SoundSpaces~\cite{chen2020soundspaces, chen2022soundspaces} is often used as a simulator in previous studies on audio-visual navigation.
There are two versions of SoundSpaces: 1.0~\cite{chen2020soundspaces} and 2.0~\cite{chen2022soundspaces}, where 
2.0 is closer to a realistic setting.
Therefore, 2.0 was used in this study.

\section{Multi-Goal Audio-Visual Navigation}

\subsection{Task Definitions}

Multi-goal audio-visual navigation proposed in this paper is an audio-visual navigation that provides navigation to multiple goals in a single episode (Fig.~\ref{fig:multi_goal_av_nav}).
Therefore, the agent must localize each sound source by observing multiple overlapping sounds.
Another major difference from MultiON, other than the use of auditory information, is that the order in which the navigation to the goal is performed is not specified.
Therefore, the agent must consider in what order it is efficient to navigate to each sound source.
In the following, a more detailed explanation of multi-goal audio-visual navigation is given.

Formally, an episode is defined by a set
$
\{E, \bm{p}_s, \theta_s, \bm{p}_{g_1}, ... , 
$
$
\bm{p}_{g_n}, S_{g_1}, ... , S_{g_n}\},
$
where $E$ represents the scene environment used in the episode, and $\bm{p}_s \in \mathbb{R}^3$ and $\theta_s \in [0, 2\pi]$ represent the starting position and direction the agent is facing, respectively.
Also, $n \in \mathbb{N}$ represents the number of goals, and for each $i \in \{1, ... , n\}$, $\bm{p}_{g_i} \in \mathbb{R}^3$ and $S_{g_i} \in SoundCategory$ represent the location of goal $g_i$ and the sound source category, respectively.
Here, $SoundCategory$ represents a set of sound categories. 
In this study, this includes, for example, \textit{telephone} and \textit{birdsong}.
A total of $91$ of sound categories are used in this study.
From the above, it can be said that multi-goal audio-visual navigation is to move from the start $(\bm{p}_s, \theta_s)$, listen to the overlapping sounds of $S_{g_1}, ..., S_{g_n}$, estimate the goal positions $\bm{p}_{g_1}, ... \bm{p}_{g_n}$, and reach them in environment $E$.

The multi-goal audio-visual navigation uses a goal format named AudioGoal~\cite{chen2020soundspaces}.
AudioGoal means that the goal is a sound source, and its location must be deduced from the information of the periodically generated sound.
Since the goal is not visually indicated, the position of the goal must be estimated based on auditory information only.

Also, when an agent reaches a goal, the sound source at that goal does not emit any sound thereafter.
In other words, once the agent reaches goal $g_i$, the agent no longer observes the sound of $S_{g_i}$.
Therefore, the sounds currently observed by the agent are emitted from sound sources that have not yet reached the goal.
The agent does not need to determine whether the sound it is currently observing is emanating from a sound source that has already arrived or from a sound source that has not yet arrived.

\subsection{Action Space}

In multi-goal audio-visual navigation, the agent's action space is
$
\{MoveForward, TurnLeft, TurnRight,
$
$
Found\}.
$
If $MoveForward$ is selected, the agent will move forward $0.25\ \mathrm{m}$ in the environment.
If $TurnLeft$ or $TurnRight$ is selected, the agent will rotate $10^{\circ}$ to the left or right, respectively.
If the agent could select a $Found$ in a radius less than $1\ \mathrm{m}$ of the goal sound source, it means that the agent has reached that goal.
The above specific values are the default settings in SoundSpaces 2.0.

An episode ends when one of the following three conditions is met.
The first is the case that all goals are reached.
The second is the case that $Found$ is selected at least $1\ \mathrm{m}$ radius away from the goal.
The third is the case that the total number of actions exceeds 2,500.
The upper limit of the number of actions was set in the previous study \cite{wani2020multion}.

\subsection{Observation Space}

The agent can observe visual information and auditory information in multi-goal audio-visual navigation.
For visual information, we use a first-person $128 \times 128$ RGBD image.
For auditory information, a $257 \times 69$ spectrogram is used.
As for the auditory information, the two-channel binaural sound is used, following previous studies.

The procedure for creating the spectrogram in this study is as follows.
First, we acquire a time series of discrete sound data for $0.25$ seconds sampled at a sampling frequency of 44,100$\ \mathrm{Hz}$.
Next, a short-time Fourier transform is performed on the time series data to obtain the amplitudes of the components of each frequency at each time.
Here, the window function is the Hanning window, a windowed signal length is $512$, and a hop length is $160$.
Finally, a spectrogram is created by adding $1$ to each value and taking the logarithm of the result as the strength of each component.

\subsection{Metrics}

Four evaluation indicators were used in this study: $SUCCESS$, $SPL$, $PROGRESS$, and $PPL$.
These are described in detail below.

The $SUCCESS$ is represented by the following, where $N$ is the number of episodes tested.
\[
SUCCESS = \frac{1}{N} \sum_{i = 1}^N S_i,
\]
where $S_i \in \{0, 1\}$ is the binary value of whether all goals were reached in the episode $i \in \{1, ..., N\}$.
That is, $S_i = 1$ if the agent was able to reach all goals in the $i$th episode, and $S_i = 0$ if the agent was unable to reach even one goal.

The $SPL$ (short for success weighted by path length) is represented by the following \cite{anderson2018evaluation}:
\[
SPL = \frac{1}{N} \sum_{i = 1}^N S_i \frac{l_i}{\max(l^A_i, l_i)}.
\]
Here, $l^A_i \in \mathbb{R}$ is the length of the path the agent took, and $l_i \in \mathbb{R}$ is the length of the shortest path to reach all goals.
In other words, even if the agent can reach all goals, $SPL$ will not be high if it does not proceed along a path that is close to the shortest path.

The $PROGRESS$ is represented by the following \cite{wani2020multion}:
\[
PROGRESS = \frac{1}{N} \sum_{i = 1}^N \frac{n^A_i}{n},
\]
where $n \in \mathbb{N}$ is the number of goals and $n^A_i \in \mathbb{N}$ is the number of goals reached by the agent in episode $i$.
In other words, unlike $SUCCESS$, the value will not be $0$ if even one goal is reached, even if not all goals are reached.

The $PPL$ (short for progress weighted by path length) is represented by the following \cite{wani2020multion}:
\[
PPL = \frac{1}{N} \sum_{i = 1}^N \frac{n^A_i}{n} \frac{l^\mathrm{MG}_i}{\max(l^A_i, l^\mathrm{MG}_i)},
\]
where $l^\mathrm{MG}_i \in \mathbb{R}$ is the length of the shortest path from the starting point to pass through all goal points reached by the agent.
In other words, $PPL$ has a higher value when more goals are reached by a path that is closer to the shortest path.
Unlike $SPL$, the value will not be $0$ if even one goal is reached, even if not all goals are reached.

However, unlike Wani et al.~\cite{wani2020multion}, the order in which goals should be reached is not determined in this study.
Therefore, the calculation method of $l^\mathrm{MG}_i$ is different.
Suppose that in episode $i$ the agents reach the goal in order $g_1, ... , g_{n^A_i}$, and the starting position is $\bm{p}_s$.
Also assume that the length of the shortest path between points $\bm{p}$ and $\bm{q}$ can be expressed as $d_{\mathrm{min}}(\bm{p}, \bm{q})$.
In this case, in Wani et al. \cite{wani2020multion}, 
\[
l^\mathrm{MG}_i = d_{\mathrm{min}}(\bm{p}_{s}, \bm{p}_{g_1}) + \sum_{j=2}^{n^A_i} d_{\mathrm{min}}(\bm{p}_{g_{j-1}}, \bm{p}_{g_j}).
\]
Since the order of goals to be reached in multi-goal audio-visual navigation task is not determined, it should be calculated as follows.
\[
l^\mathrm{MG}_i = \min_{\sigma \in T_{n^A_i}} \!\left\{\! d_{\mathrm{min}}(\bm{p}_{s}, \bm{p}_{g_{\sigma(1)}}) +  \sum_{j=2}^{n^A_i} d_{\mathrm{min}}(\bm{p}_{g_{\sigma(j-1)}}, \bm{p}_{g_{\sigma(j)}}) \!\right\}
\]
where $T_{n^A_i}$ is the entire set of permutations of $n^A_i$ numbers $1, 2, ... ,n^A_i$.
\section{METHODS}

\subsection{Baselines}

We will test two well-known deep reinforcement learning methods as baseline methods, in addition to the simplest random action selection method.
Each method is described in detail below.

\subsubsection{Random}
This agent randomly chooses an action from among $\{MoveForward, TurnLeft, TurnRight\}$.
However, if the agent is within a radius of less than $1\ \mathrm{m}$ of the goal sound source, it will always choose $Found$.

\subsubsection{AV-Nav \cite{chen2020soundspaces}}
An end-to-end deep reinforcement learning method.
It has a GRU-based policy network with first-person images and spectrograms as input.
It is the first and simplest method proposed for audio-visual navigation tasks.

\subsubsection{SAVi \cite{chen2021semantic}}
A deep reinforcement learning method with a Transformer-based policy network.
In addition to first-person images and spectrograms, the agent's position, orientation, and previous actions are used as input.
Also, it has a goal descriptor network that predicts the category and the location of the goal sound source. The category prediction part is pre-trained.
The learning of policy network is divided into two stages.
In the first stage, the memory size of the Transformer is set to 1 and the observation embedding is learned.
In the second stage, the learning of the observation embedding is frozen and the memory size is increased to 150 to learn the rest of the network.

Since the target task of this study is not semantic audio-visual navigation~\cite{chen2021semantic} and also due to the difficulty of extending to multiple goals, the goal descriptor network was not used in this study\footnote{We actually tested SAVi with the goal descriptor network, but the performance was degraded.}.

\subsection{Proposed Method}


\begin{figure*}[t]
    \begin{center}
        \centering
        \includegraphics[scale=0.6]{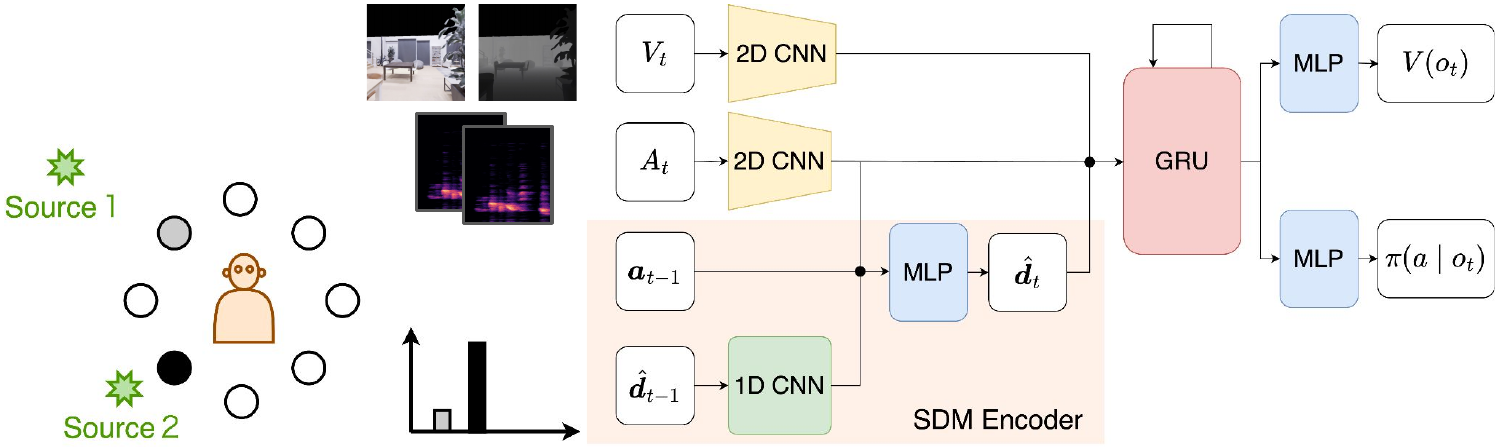}
        \caption{
            The application of the network for SDM creation to the network architecture of the AV-Nav \cite{chen2020soundspaces}.
            SDM Encoder is the proposed network architecture for SDM creation.
            Here, $\boldsymbol{a}_{t}$ and $\hat{\boldsymbol{d}}_t$ represent one-hot vector representing action and SDM prediction at time $t$, respectively.
            The input includes not only the sound observation $A_t$ but also one-hot vector $\boldsymbol{a}_{t-1}$ representing the previous action and the prediction of the previous SDM $\hat{\boldsymbol{d}}_{t-1}$.
        }
        \label{fig:direct_map_encoder}
    \end{center}
\end{figure*}

In the following, we describe a proposed method, sound direction map (SDM), which allows for explicit and dynamic localization of multiple sound sources using memory.
The SDM is a representation of how far and in which direction the sound source is located from the agent.
On the left of Fig.~\ref{fig:direct_map_encoder}, SDM is represented by black nodes surrounding the agent.
This is an example of a case with two sound sources.
In SDM, the closer the distance from the agent to the sound source, the higher the value of the node in the direction of that sound source.
The color of the node represents the increase in the value of the SDM node.
In this way, SDM allows localization by direction and distance for multiple sound sources.
However, if there are multiple sound sources in the same direction, only the closest sound source is considered.
If the agent is unable to separate the sound sources, accurately selecting the closer one and predicting the distance may be difficult for the agent.

The value of each node is the reciprocal of the geodesic distance to the sound source in that direction.
The reason for using geodesic distance rather than Euclidean distance is that sound waves are reflected, diffracted, and attenuated by walls and other obstacles. Geodesic distance is therefore considered to be more predictable.
In addition, clipping was used for sound sources within $1\ \mathrm{m}$ of the geodesic distance.
There are two reasons for this. One is that the agents do not need to perform strict localization as long as they are within a radius of $1\ \mathrm{m}$.
The other is to avoid overreacting to very large values when training the encoders that create the SDM.

We propose a method that uses a neural network to dynamically predict SDM while the agent repeats its actions.
In this study, SDMs were applied to the AV-Nav \cite{chen2020soundspaces} and SAVi \cite{chen2021semantic} networks.
The whole network is trained in an end-to-end manner.
The proposed neural network architecture when AV-Nav~\cite{chen2020soundspaces} is used as the backbone is shown in Fig.~\ref{fig:direct_map_encoder}.
The current audio observation, the previous action, and the previous SDM are used to predict the current SDM. The current audio observation is necessary to predict the location of sound sources.
In SAVi, we have added SDM encoder as part of the observation encoder.
Therefore, in SAVi, SDM encoder is trained only in the first stage.
In addition to the gradients flowing from the policy networks, the gradients from the Mean Squared Error (MSE) between the predicted SDM and the true SDM were used to update the weights of the SDM encoder.

Also, when training, the previous SDM's true value $\boldsymbol{d}_{t-1}$ is input as the previous SDM.
In our preliminary experiments, we also tested the case in which the previous own prediction $\hat{\boldsymbol{d}}_{t-1}$ was input as the previous SDM for training.
Experiments showed that learning by inputting true values rather than predictions performed better, so that method was used.
However, when testing, we did not use the true value but the prediction value.
We also used the prediction value for the second stage of SAVi training.

In addition, Dropout was used during training.
This means that the value of each node in the SDM that is input as the previous SDM is set to 0 with a certain probability.
The purpose of this is to prevent too much reliance on the previous SDM when predicting the current SDM.
This is because the predicted previous SDM $\hat{\boldsymbol{d}}_{t-1}$ can fail to be close to the true previous SDM $\boldsymbol{d}_{t-1}$ properly.
Note that Dropout is not performed in the SAVi's second stage of training.

\subsection{Reward}

The reward received by the agent at time $t$ is defined as
$
r_t = r_{\mathrm{found}} - \Delta_{\mathrm{geo}} - 0.01, \label{eq:reward_definition}
$
where $r_{\mathrm{found}}$ is $5$ if the agent reached the goal at time $t$ and $0$ otherwise.
Also, $\Delta_{\mathrm{geo}} \in \mathbb{R}$ is the  change of minimum geodesic distance to reach all goals that have not yet been reached.
In other words, $\Delta_{\mathrm{geo}}$ is negative when the minimum geodesic distance to reach all goals not yet reached becomes small and positive when it becomes large.

\subsection{Training}

The agents were trained using a distributed deep reinforcement learning method called decentralized distributed proximal policy optimization (DD-PPO) \cite{wijmans2019dd}.
For training with AV-Nav, 4 GPUs were used and four workers were placed on each GPU.
For training with SAVi, 16 GPUs were used and two workers were placed on each GPU.
The SAVi was found to take longer to train than AV-Nav, so the hardware set-up was modified in this way.

The structure of 1D CNN and MLP in SDM Encoder is as follows.
First, the 1D CNN has four layers.
The number of output channels is 32, the size of the kernel is 3, circular padding is used for padding, and ReLU is used as the activation function.
Second, MLP has 4 layers.
The output sizes are 1048, 1048, 524, and 8 in this order.
For the activation function, a sigmoid function is used only for the last layer, and ReLU is used for the other layers.

The hyperparameters in this study were set as follows.
First, the number of SDM nodes was set to $8$, and the probability of Dropout being performed in each node was set to $0.2$.
Furthermore, in the gradient when updating the weights of the SDM encoder, the coefficient of the gradient due to the MSE between the predicted SDM and the true SDM is set to $100$.
In preliminary experiments, 1, 10, 100, and 1000 were tried, and 100 was adopted because it gave the best performance.
In addition, the default parameters of SoundSpaces were used for the hyperparameters related to AV-Nav and SAVi.
The number of times the weights are updated is 1,250 times for AV-Nav, while for SAVi the first stage is 1,190 times and the second stage is 810 times.

\section{Experiments}

\subsection{Implementation Details}

\subsubsection{Simulation}
Agents were trained and tested on a simulator named SoundSpaces \cite{chen2020soundspaces, chen2022soundspaces} and a scene dataset named Replica \cite{straub2019replica}.
SoundSpaces 2.0 \cite{chen2022soundspaces} used in this study is a simulator that extends a visual rendering simulator Habitat-Sim \cite{savva2019habitat} by integrating an acoustic propagation engine RLR-Audio-Propagation. 
Replica is a scene dataset consisting of 18 different apartment, office, room, and hotel scenes.

Scenes for training, validation, and testing are divided in the same way as in the previous study \cite{chen2020soundspaces}.
Therefore, scenes not used during training are used during testing.
Also, we did not use the dataset for validation but tested with the parameters obtained from the last parameter update.
Here, the number of test episodes is 1,000 for all experiments.

\subsubsection{Episode generation}

In this study, $\boldsymbol{p}_s, \theta_s, \boldsymbol{p}_{g_1}, ... , \boldsymbol{p}_{g_n}$ are subject to some constraints when they are generated to eliminate episodes that are too easy and too difficult.

First, to eliminate episodes that are too easy, the distances between each point $\boldsymbol{p}_s, \boldsymbol{p}_{g_1}, ..., \boldsymbol{p}_{g_n}$ are to be at least $1\ \mathrm{m}$ apart and the ratio of the geodesic distance to the Euclidean distance is to be greater than $1.1$.
The second reason for the constraint is to eliminate cases where the goal can be reached almost exclusively in a straight line.
However, since it was difficult to satisfy these constraints in room 2 and office 1 due to their narrowness, we decided that the distance between each point should be at least $0.6\ \mathrm{m}$, and the ratio of the geodesic distance to the Euclidean distance should be greater than $1.001$ in these.

To eliminate episodes that are too difficult, the height between each point was made to be less than $0.3\ \mathrm{m}$, and the distance (m) between each point $d$ was made to be less likely to increase in apartment 0 due to its wideness.
Specifically, the locations are rejected with probability $p=1.0$ if $d>10$, with $p=0.7$ if $d>6$, with $p=0.6$ if $d>5$, with $p=0.5$ if $d>4$, with $p=0.4$ if $d>3$, and with $p=0$ if $d<3$.
The reason for this constraint is that we found that without this constraint, the performance in only apartment 0 would be significantly lower and the learning curve would be unstable.
We believe that the essential solution to this problem requires the introduction of curriculum learning.

\subsubsection{Sound sources}
Unless otherwise noted, we use 73 different sound sources for training and 18 different sound sources for testing, following the previous study \cite{chen2020soundspaces}.
Here, there are no sound sources that overlap between training and testing.
Therefore, the test evaluates generalization performance for sounds that were never heard during training.

All sound source is $1$ second of sound data sampled at a sampling frequency of 44,100$\ \mathrm{Hz}$.
This sound data is played repeatedly until the agent reaches its sound source.
Unless otherwise noted, in each episode, the sound data is made to start playing at a random time between $0$ and $1$ seconds.

\subsection{Comparison by the number of goals}
\label{comparison-by-the-number-of-goals}
First, an experiment was conducted using the baseline methods, varying only the number of goals.
The purpose of this experiment was to investigate the differences in difficulty with the number of goals.
The number of goals $n$ was performed in three ways: $n=1,2,3$.
The test results are shown in TABLE \ref{tab:n_goal_results}.

\begin{table}[tb]
    \setlength{\tabcolsep}{4pt}
    \centering
    \caption{
        Comparison by the number of goals.
        Here, the number of goals is $n$, meaning that there are $n$ goals in all episodes of training and testing.
    }
    \label{tab:n_goal_results}
    \begin{tabular}{@{}cccccc@{}}
    \toprule
        Method & $n$ & $SUCCESS$ & $SPL$ & $PROGRESS$ & $PPL$ \\ \midrule
        Random & 1 & \textbf{0.432} & \textbf{0.141} & \textbf{0.432} & \textbf{0.141} \\
        & 2 & 0.167 & 0.048 & 0.377 & 0.070 \\
        & 3 & 0.053 & 0.017 & 0.317 & 0.055 \\ \midrule
        AV-Nav~\cite{chen2020soundspaces} & 1 & \textbf{0.503} & \textbf{0.323} & \textbf{0.503} & \textbf{0.323} \\
        & 2 & 0.179 & 0.119 & 0.229 & 0.142 \\
        & 3 & 0.107 & 0.071 & 0.292 & 0.160 \\ \midrule
        SAVi~\cite{chen2021semantic} & 1 & \textbf{0.771} & \textbf{0.507} & \textbf{0.771} & \textbf{0.507} \\
        & 2 & 0.643 & 0.416 & 0.720 & 0.438 \\
        & 3 & 0.226 & 0.138 & 0.449 & 0.215 \\ \bottomrule
        \end{tabular}
\end{table}

We found that increasing the number of goals tends to cause a large drop in accuracy.
We believe there are two reasons for the large drop in $SUCCESS$.
The first reason is that $SUCCESS$ degrades exponentially.
This is because if the probability of reaching one goal from the start is $p$, the probability of reaching $n$ goals is $p^n$.
The second reason is that the need for sound source separation arises, which will be discussed in more detail in Section~\ref{invest_difficulties}.
We also believe that the reason for the lower $PROGRESS$ is due to the end condition of the episode.
In this setting, if navigating to a goal in the middle of the episode fails, the episode will end.
Thus, if there are multiple goals remaining, the failure of navigating to one goal will fail to navigate to multiple goals.
This is considered to have lowered $PROGRESS$ as the reachability is lowered.

\begin{figure*}[t]
    \begin{center}
        \centering
        \includegraphics[scale=0.7]{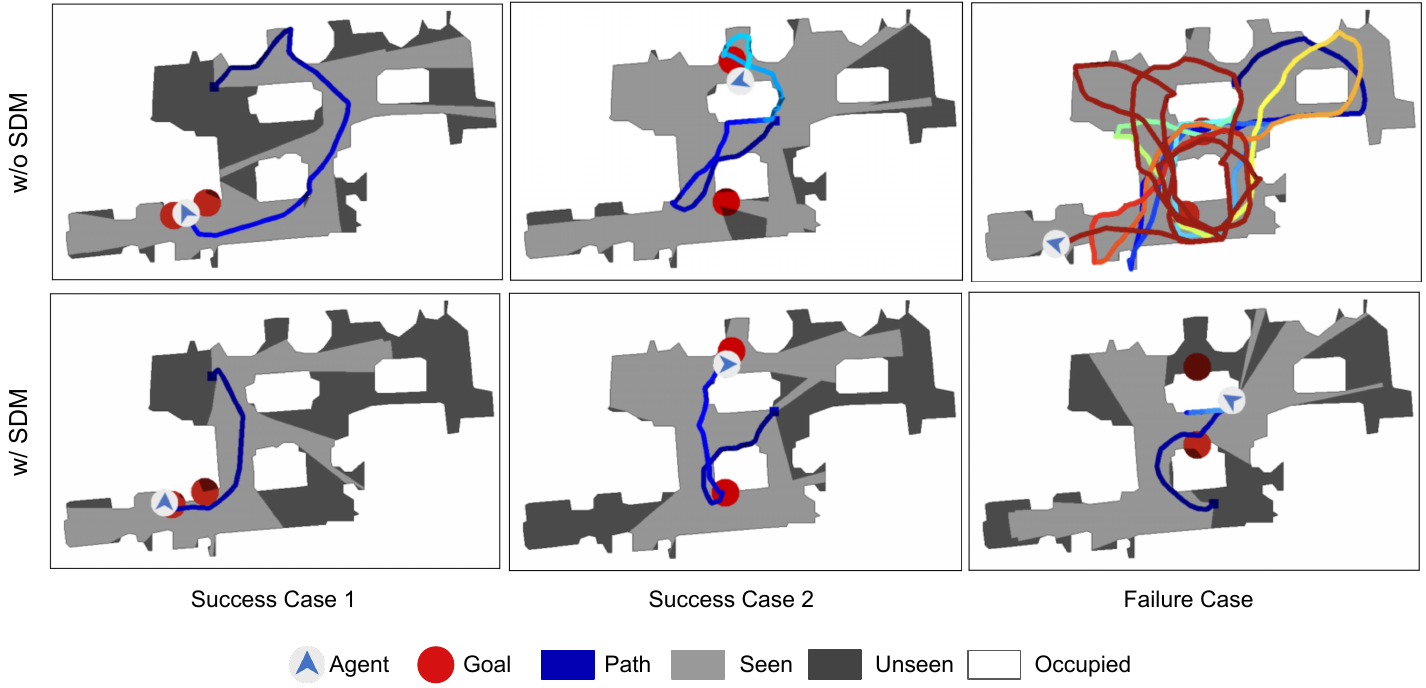}
        \caption{
            Navigation trajectory comparison.
            The upper row is AV-Nav w/o SDM and the lower row is AV-Nav w/ SDM.
            The color of the Path represents the step elapsed.
            It changes from blue to red as the steps elapse.
            It can be seen that the use of SDM has reduced the number of unnecessary actions.
        }
        \label{fig:qualitative_eval}
    \end{center}
\end{figure*}

\subsection{Investigating difficulties with multiple sound source goals}
\label{invest_difficulties}

We investigate the difficulties that lie in multi-goal audio-visual navigation.
In this section, we used AV-Nav~\cite{chen2020soundspaces} for the navigation performance evaluation.

\subsubsection{Loud and quiet sounds}

We investigated whether the difficulty level varies with the loudness of the sound, i.e., with the volume of the sound.
Here, we compare the difference in reachability to quiet and loud sounds.
The ratio of the number of goals reached to the number of all goals is shown here. In other words, if the goals are selected from the same set, it is the same as $PROGRESS$. However, if the goals are selected from different sets, it is slightly different from $PROGRESS$.
All of these sounds were selected from those not used in the training.
The sounds were selected so that the average length of the sounds in the quiet sound set and the loud sound set are close to each other.

\begin{table}[tb]
    \caption{
        Comparison of reachability to quiet (top) and loud (bottom) sounds.
        $n$-quiet and $n$-loud means that $n$ goals were selected from the set of quiet sounds and the set of loud sounds, respectively.
        Note that all 73 sound sources for training were used during training.
    }
    \label{tab:loud_and_quiet}
    \centering
    \begin{tabular}{@{}ccccccc@{}}
      \toprule
      & \multicolumn{2}{c}{1-goal task} & \multicolumn{3}{c}{2-goal task} \\
      \cmidrule(lr){2-3} \cmidrule(lr){4-6}
      & 1-quiet & 1-loud & 2-quiet & 2-loud & 1-quiet-1-loud \\ \midrule
      quiet & 0.441 & N/A & 0.232 & N/A & 0.187 \\
      loud & N/A & 0.449 & N/A & 0.209 & 0.253 \\
      \bottomrule
    \end{tabular}
\end{table}

The results are shown in TABLE \ref{tab:loud_and_quiet}.
It shows that the accuracy of both loud and quiet sounds decreases when a loud sound is heard at the same time.
We believe this is because loud sounds can be heard at a distance and therefore tend to become noise even if the other sound is also loud.

\subsubsection{Long and short sounds}

We investigated whether the difficulty level varies with the length of time a sound is played.
Here we compare the difference in reachability to short and long sounds.
As in the previous section, the ratio of the number of goals reached to the number of all goals is shown here. 
All of these sounds were selected from those not used in training.
Also, these sounds were selected so that the average of the maximum volume of the short sound set and long sound set are close to each other.

\begin{table}[tb]
    \centering
    \caption{
        Comparison of reachability to short (top) and long (bottom) sounds.
        $n$-short and $n$-long mean that $n$ goals were selected from the set of short sounds and the set of long sounds, respectively.
        Note that all 73 sound sources for training were used during training.
    }
    \label{tab:long_and_short}
    \begin{tabular}{@{}ccccccc@{}}
    \toprule
      & \multicolumn{2}{c}{1-goal task} & \multicolumn{3}{c}{2-goal task} \\
      \cmidrule(lr){2-3} \cmidrule(lr){4-6}
      & 1-short & 1-long & 2-short & 2-long & 1-short-1-long \\ \midrule
      short & 0.235 & N/A & 0.160 & N/A & 0.141 \\
      long & N/A & 0.632 & N/A & 0.255 & 0.262 \\ \bottomrule
    \end{tabular}
\end{table}

The results are shown in TABLE \ref{tab:long_and_short}.
It shows that the accuracy of both long and short sounds decreases when a long sound is played at the same time.
We believe that this is because long sounds are always being played and thus tend to become noise for the other sound.

\subsubsection{Same and different sounds}

We investigated whether the difficulty level varied depending on whether multiple sound types were the same or different.
Here, "same" 
means the same sound data is used.
Also, two types of sound sources were used randomly, allowing duplicates during training. Three situations were performed during testing: two same types, two different types, and two random types allowing duplicates.

\begin{table}[tb]
    \centering
    \caption{
        Comparison in different situations where there are two same, different, and random sounds.
    }
    \label{tab:same_and_different_sound}
    \begin{tabular}{@{}ccccc@{}}
    \toprule
        & $SUCCESS$ & $SPL$ & $PROGRESS$ & $PPL$ \\ \midrule
        same & \textbf{0.193} & \textbf{0.131} & \textbf{0.244} & \textbf{0.155} \\
        different & 0.181 & 0.120 & 0.232 & 0.145 \\
        random & 0.179 & 0.119 & 0.229 & 0.142  \\ \bottomrule
        \end{tabular}
\end{table}

The results are shown in TABLE \ref{tab:same_and_different_sound}.
It was found that accuracy was lower when different sounds were sounding.
We suspect that this is because sounds of different natures reduce the reachability of one sound, just as sounding a loud sound and a quiet sound reduces the reachability of a quiet sound, and sounding a long sound and a short sound reduces the reachability of a short sound when the two sounds are different.
We suspect that when they are different, one is drowned out by the other, thus lowering the reachability.
These results indicate the importance of sound source separation.

\subsubsection{Timing of sounding}

\begin{table}[tb]
    \centering
    \caption{
        Comparison by different timing of two goal sounding.
        Only \textit{telephone} sound was used in this experiment.
        Also, the training is done in a random setting.
    }
    \label{tab:sound_timing}
    \begin{tabular}{@{}ccccc@{}}
    \toprule
        & $SUCCESS$ & $SPL$ & $PROGRESS$ & $PPL$ \\ \midrule
        overlap & \textbf{0.792} & \textbf{0.430} & \textbf{0.868} & \textbf{0.444} \\
        non-overlapping & 0.735 & 0.410 & 0.811 & 0.424 \\
        random & 0.736 & 0.413 & 0.820 & 0.440 \\ \bottomrule
    \end{tabular}
\end{table}

We investigated whether the difficulty varied depending on the timing of multiple sounds sounding.
The results are shown in TABLE \ref{tab:sound_timing}.
When two identical sounds are sounding, the result is that accuracy is lower when there is no overlap between the sounds.
We believe that this is because it is more difficult to localize the sound source when the same sound is played alternately.
Since the number of sound sources is not given to the agent, if the same sound is heard in different places alternately, it may be judged that one sound is coming and going.
Therefore, a method to represent the history of dynamic multiple sound source localization is considered to be important.

\subsection{Sound Direction Map}

\subsubsection{Quantitative evaluation}

To demonstrate the usefulness of SDM, we compared the performance with and without SDM for two baselines.
The results are shown in TABLE \ref{tab:dm_results}.
Here, the experimental procedure is the same as in Section \ref{comparison-by-the-number-of-goals}.
For all the number of goals and all baselines, performance was improved by using SDM.
Furthermore, we found that SDM tended to suppress the degradation caused by an increase in the number of goals.
We believe that the reason is that SDM effectively utilizes memory and allows for more accurate localization for multiple sound sources.

However, these results did not reveal whether there is a limit on the number of achievable goals. The number of goals the agent reached, expressed as $n \times PROGRESS$, still tends to increase when calculated based on the results in TABLE \ref{tab:dm_results}.

\begin{table}[tb]
    \setlength{\tabcolsep}{3pt}
    \centering
    \caption{
        Quantitative evaluation of SDM.
        $n$ denotes the number of goals.
    }
    \label{tab:dm_results}
    \begin{tabular}{@{}llcccc@{}}
    \toprule
        $n$~ & method & $SUCCESS$ & $SPL$ & $PROGRESS$ & $PPL$ \\ \midrule
        1 & AV-Nav~\cite{chen2020soundspaces} & 0.503 & 0.323 & 0.503 & 0.323  \\
        & SAVi~\cite{chen2021semantic} & 0.771 & 0.507 & 0.771 & 0.507 \\
        & AV-Nav~\cite{chen2020soundspaces} w/ SDM & 0.610 & 0.354 & 0.610 & 0.354 \\
        & SAVi~\cite{chen2021semantic} w/ SDM & \textbf{0.838} & \textbf{0.616} & \textbf{0.838} & \textbf{0.616} \\ \midrule
        2 & AV-Nav~\cite{chen2020soundspaces} & 0.179 & 0.119 & 0.229 & 0.142  \\
        & SAVi~\cite{chen2021semantic} & 0.643 & 0.416 & 0.720 & 0.438 \\
        & AV-Nav~\cite{chen2020soundspaces} w/ SDM & 0.332 & 0.172 & 0.506 & 0.232 \\
        & SAVi~\cite{chen2021semantic} w/ SDM & \textbf{0.764} & \textbf{0.464} & \textbf{0.822} & \textbf{0.480} \\ \midrule
        3 & AV-Nav~\cite{chen2020soundspaces} & 0.107 & 0.071 & 0.292 & 0.160 \\
        & SAVi~\cite{chen2021semantic} & 0.226 & 0.138 & 0.449 & 0.215 \\
        & AV-Nav~\cite{chen2020soundspaces} w/ SDM & 0.174 & 0.101 & 0.368 & 0.186 \\
        & SAVi~\cite{chen2021semantic} w/ SDM & \textbf{0.469} & \textbf{0.319} & \textbf{0.615} & \textbf{0.385} \\ \bottomrule
    \end{tabular}
\end{table}

\subsubsection{Qualitative evaluation}

Fig.~\ref{fig:qualitative_eval} compares the trajectories of the agents with and without SDM.
It can be seen that the use of SDM has reduced the number of unnecessary actions.
We believe that this is due to more accurate sound source localization.
In addition, in the cases where AV-Nav w/o SDM failed, there were examples of long wandering.
We believe this may be due to inaccurate sound source localization and the inability to determine where the goal is located.
Also, without SDM, there were cases of failure due to being caught by obstacles.
We believe this is because the SDM has determined that the sound source is on the opposite side of the obstacle.

\section{CONCLUSION}
\label{CONCLUSION}

In this paper, we proposed a new framework \textit{multi-goal audio-visual navigation}, in which multiple sound sources serve as goals. We investigated the effect of increasing the number of goals on navigation performance.
The results showed that increasing the number of goals tends to cause large performance degradation.
In particular, we found that navigation became difficult when there is a loud or long sound.
Subsequent difficulty investigations implied the importance of sound source separation and also suggested the importance of effective use of memory.

We also proposed a method for solving the multi-goal audio-visual navigation task with higher accuracy, sound direction map (SDM).
The SDM takes advantage of memory to dynamically localize multiple sound sources.
Experiments showed that the SDM is useful for all the numbers of goals and all baselines.
Qualitative results also demonstrated that the SDM successfully decreased the number of unnecessary actions.

\addtolength{\textheight}{-12cm}   







\bibliographystyle{junsrt}
\bibliography{references}

\end{document}